\documentclass[10pt, conference, compsocconf]{IEEEtran}
\usepackage{times}
\usepackage{url}
\usepackage{latexsym}

\usepackage{caption}
\usepackage{graphicx}
\usepackage{color}
\usepackage{amsmath}
\usepackage{amssymb} 
\usepackage{booktabs}
\usepackage[normalem]{ulem}
\usepackage{comment}
\usepackage{multirow}
\usepackage{subfigure}
\usepackage{balance}

\usepackage{float}
\restylefloat{table}

\DeclareMathOperator \real{\mathbb{R}}

\newcommand{\alert}[1]{\textcolor{red}{\noindent$\Rightarrow$ #1}}

\DeclareMathOperator*{\Ber}{Ber}
\DeclareMathOperator*{\sig}{sig}

\newcommand{\Ni}{({\em i})~}
\newcommand{\Nii}{({\em ii})~}
\newcommand{\Niii}{({\em iii})~}

\usepackage{placeins}

\ifCLASSINFOpdf
\else
\fi

\begin{document}
%
\title{Rapid Classification of Crisis-Related Data on Social Networks using\\ Convolutional Neural Networks}


\author{\IEEEauthorblockN{Dat Tien Nguyen, Kamela Ali Al Mannai, Shafiq Joty, Hassan Sajjad, Muhammad Imran, Prasenjit Mitra}
\IEEEauthorblockA{Qatar Computing Research Institute -- HBKU, Qatar\\
\{ndat, kamlmannai, sjoty, hsajjad, mimran, pmitra\}@qf.org.qa}
}


%


\maketitle

\begin{abstract}
The role of social media, in particular microblogging platforms such as Twitter, as a conduit for actionable and tactical information during disasters is increasingly acknowledged. However, time-critical analysis of big crisis data on social media streams brings challenges to machine learning techniques, especially the ones that use supervised learning. Scarcity of labeled data, particularly in the early hours of a crisis, delays the machine learning process. The current state-of-the-art classification methods require a significant amount of labeled data specific to a particular event for training plus a lot of feature engineering to achieve best results. In this work, we introduce neural network based classification methods for binary and multi-class tweet classification task. 
We show that neural network based models do not require any feature engineering and perform better than state-of-the-art methods. In the early hours of a disaster when no labeled data is available, our proposed method makes the best use of the out-of-event data and achieves good results.

\end{abstract}

\begin{IEEEkeywords}
Deep learning, Neural networks, Supervised classification, Social media, Crisis response 

\end{IEEEkeywords}

%

\section{Introduction}

Time-critical analysis of social media data streams is important for many application areas~\cite{lee2013real,o2010tweets,rudra2016summarizing}. During the onset of a crisis situation, people use social media platforms to post situational updates, look for useful information, and ask for help~\cite{castillo2016big,imran2015processing}. Rapid analysis of messages posted on microblogging platforms such as Twitter can help humanitarian organizations like the United Nations gain situational awareness, learn about urgent needs of affected people, critical infrastructure damage, medical emergencies, etc. at different locations, and  decide on response actions accordingly~\cite{acar2011twitter,vieweg2014integrating}. 

Artificial Intelligence for Disaster Response\footnote{http://aidr.qcri.org/}  (AIDR) is an online platform to support disaster response utilizing big crisis data from social networks~\cite{imran2014aidr}. During a disaster, any person or organization can provide keywords 
to collect tweets of interest. 
Filtering using keywords helps cut down the volume of the data to some extent. But, identifying different kinds of {\it useful} tweets that responders can act upon cannot be achieved using only keywords because a large number of tweets may contain the keywords but are of limited utility for the responders. The best-known solution to address this problem is to use supervised classifiers that would separate useful tweets from the rest and classify them accordingly into informative classes.


Automatic classification of tweets to identify useful tweets is a challenging task because: \Ni  tweets are short -- only 140 characters -- and therefore, hard to understand without enough context; \Nii they often contain abbreviations, informal language, spelling variations and are ambiguous; and, \Niii judging a tweet's utility is a subjective exercise. 
Individuals differ on their judgment about whether a tweet is useful or not. 
Classifying tweets into different topical classes posits further difficulties because: 
\Ni tweets may actually contain information that belongs to multiple classes; choosing the dominant topical class is hard; and, \Nii the semantic ambiguity in the language used sometimes makes it hard to interpret the tweets.
Given these inherent difficulties, a computer cannot generally 
agree with annotators at a rate that is higher than the rate at which the annotators agree with each other.
Despite advances in natural language processing (NLP), interpreting the semantics of the short informal texts automatically remains a hard problem. 




To classify crisis-related data, traditional classification approaches use batch learning with discrete representation of words. This approach has three major limitations. First, in the beginning of a disaster situation, there is no event labeled data available for training. Later, the labeled data arrives in small batches depending on the availability of geographically dispersed volunteers (e.g. in case of AIDR). 
These learning algorithms are dependent on the labeled data of the event for training. Due to the discrete word representations and the variety across events from which the big crisis data is collected, they perform poorly
when trained on the data from previous events (\emph{out-of-event data}). 
Second, training a classifier from scratch every time a new batch of labeled data arrives is
infeasible due to the velocity of the big crisis data.
Third, traditional approaches require manually engineered features like cue words and TF-IDF vectors \cite{imran2015processing} for learning.
Due to the variability of the big crisis data, adapting the model to changes in features and their importance manually is 
undesirable (and often infeasible).

Deep neural networks (DNNs) are ideally suited for big crisis data.
They are usually trained using
online learning and have the flexibility to adaptively learn from new batches of labeled data without requiring to retrain from scratch. Due to their distributed word representation, 
they generalize well and make better use of the previously labeled data from other events to speed up the classification process in the beginning of a disaster. DNNs
obviate the need of manually crafting features and 
automatically learn latent features as distributed dense vectors, which generalize well and have shown to benefit various NLP tasks \cite{collobert2011natural,Bengio03,mikolov2013distributed,socher2013recursive}. 

In this paper, we propose a
convolutional neural network (CNN) for the classification tasks. CNN captures the most salient $n$-gram information by means of its convolution and max-pooling operations. On top of the typical CNN, we propose an extension that combines multilayer perceptron with a CNN. 
%
We present a series of experiments using different variations of the training data -- event data only, out-of-event data only and a concatenation of both. Experiments are conducted for binary and multi-class classification tasks. 
For the event only binary classification task, the CNN model outperformed in all tasks with an accuracy gain of up to 7.5 absolute points over the best non-neural classification algorithm. In the scenario of no event data, the CNN model shows substantial improvement of up to 10 absolute points over several non-neural models. 
This makes the neural network model an ideal choice for classification in 
the early hours of a disaster when no labeled data is available. When combined the event data with the out-of-event data,
overall results of all classification methods 
drop by a few points compared to 
using event only training. 

For multi-class classification, we achieved similar results as in the case of binary classification. The CNN model outperformed other competitors. 
Our variation of the CNN model with multilayer perceptron (MLP-CNN) performed better than its 
CNN-only counter part. Similar to binary classification, 
blindly adding out-of-event data either drops the performance or does not give any noticeable improvement over the event only model. 
To reduce the negative effect of large out-of-event data and to make the most out of it, we apply two simple domain adaptation techniques -- 1) weight the out-of-event labeled tweets based on their closeness to the event data, 2) select a subset of the out-of-event labeled tweets that are correctly labeled by the event-based classifier. Our results show that later results in better classification model. 


To summarize, we show that neural network models perform better than non-neural models for both binary and multi-classification scenario. They can be used reliably with the already available out-of-event data for binary and multi-class classification of disaster-related big crisis data. 
DNNs are better suited for big crisis data than conventional classifiers because they can learn features automatically and can be adopted to online settings. 

The rest of the paper is organized as follows.  We present the convolutional neural model for classification tasks in Section \ref{sec:CNN}. Section \ref{sec:adap} presents the adaptation method. We describe the dataset and training settings of the models 
in Section~\ref{sec:settings}. In Section~\ref{sec:results}, we present our results and analysis. We provide an in-depth discussion on the obtained results in Section~\ref{sec:discussion}. We summarize related work in Section \ref{sec:relatedwork}. We conclude and discuss future work in Section~\ref{sec:conclusion}.

\section{Convolutional Neural Network} \label{sec:CNN}

\begin{figure}[tb!]
\centering
\includegraphics[height=2.5in]{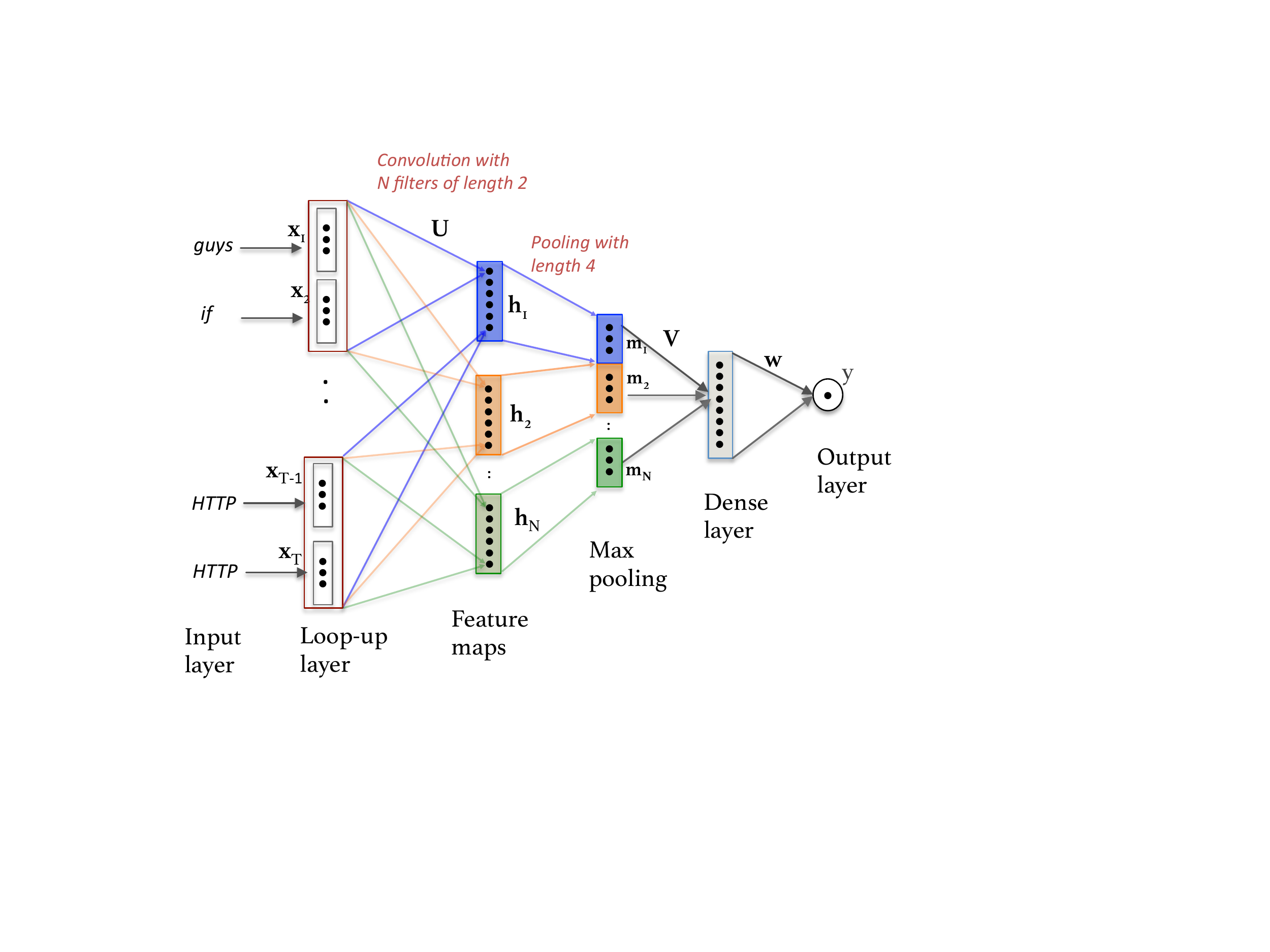}
\caption{Convolutional neural network on a tweet.}
\label{fig:cnn}
\end{figure}

When there is not enough in-event training data available, 
in order to classify short and noisy big crisis data effectively, a classification model should use 
a distributed representation of words.
Such a representation results in improved generalization.
The system should learn  the key features at different levels of abstraction automatically. 
To this end, we use a Convolutional Neural Network (CNN). 
Figure \ref{fig:cnn} demonstrates how 
a CNN works with an example tweet ``guys if know any medical emergency around balaju area you can reach umesh HTTP doctor at HTTP HTTP''.\footnote{HTTP represents URLs present in a tweet.}
Each word in the vocabulary $V$ is represented by a $D$ dimensional vector in a shared look-up table $L$ $\in$ $\real^{|V| \times D}$. $L$ is considered a model parameter to be learned. We can initialize $L$ randomly or
using pretrained word embedding vectors like word2vec \cite{mikolov2013efficient}.

Given an input tweet $\mathbf{s} = (w_1, \cdots, w_T)$, we first transform it into a feature sequence by mapping each word token $w_t \in \mathbf{s}$ to an index in $L$. The look-up layer then creates an input vector $\mathbf{x_t}\in\real^{D}$ for each token $w_t$, which are passed through a sequence of convolution and pooling operations to learn high-level feature representations. 

A convolution operation involves applying a \emph{filter} $\mathbf{u} \in \real^{L.D}$ (i.e., a vector of parameters) to a window of $L$ words to produce a new feature  


\begin{equation}
h_t = f(\mathbf{u} . \mathbf{x}_{t:t+L-1} + b_t)
\end{equation}

\noindent where $\mathbf{x}_{t:t+L-1}$ denotes the concatenation of $L$ input vectors, $b_t$ is a bias term, and $f$ is a nonlinear activation function (e.g., $\sig, \tanh$). A filter is also known as a kernel or feature detector. We apply this filter to each possible $L$-word window in the tweet to generate a \emph{feature map} $\mathbf{h}^i = [h_1, \cdots, h_{T+L-1}]$. We repeat this process $N$ times with $N$ different filters to get $N$ different feature maps (i.e., $i= 1 \cdots N$). We use a \emph{wide} convolution \cite{Kalchbrenner14} (as opposed to \emph{narrow}), which ensures that the filters reach the entire sentence, including the boundary words. This is done by performing \emph{zero-padding}, where out-of-range (i.e., $t$$<$$1$ or $t$$>$$T$) vectors are assumed to be zero.   

After the convolution, we apply a max-pooling operation to each feature map.

\begin{equation}
\mathbf{m} = [\mu_p(\mathbf{h}^1), \cdots, \mu_p(\mathbf{h}^N)] \label{max_pool}
\end{equation}
  
\noindent where $\mu_p(\mathbf{h}^i)$ refers to the $\max$ operation applied to each window of $p$ features in the feature map $\mathbf{h}^i$. For instance, with $p=2$, this pooling gives the same number of features as in the feature map (because of the zero-padding).
Intuitively, the filters compose local $n$-grams into higher-level representations in the feature maps, and max-pooling reduces the output dimensionality while keeping the most important aspects from each feature map.


Since each convolution-pooling operation is performed independently, the features extracted become invariant in locations (i.e., where they occur in the tweet), thus acts like bag-of-$n$-grams. However, keeping the \emph{order} information could be important for modeling sentences. In order to model interactions between the features picked up by the filters and the pooling, we include a \emph{dense} layer of hidden nodes on top of the pooling layer 

\begin{equation}
\mathbf{z} = f(V\mathbf{m} + \mathbf{b_h}) \label{dense} 
\end{equation}

\noindent where $V$ is the weight matrix, $\mathbf{b_h}$ is a bias vector, and $f$ is a non-linear activation. The dense layer naturally deals with variable sentence lengths by producing fixed size output vectors $\mathbf{z}$, which are fed to the output layer for classification. 

Depending on the classification task, the output layer defines a probability distribution. For binary classification task, it defines a Bernoulli distribution:  

\begin{equation}
p(y|\mathbf{s}, \theta)= \Ber(y| \sig(\mathbf{w^T} \mathbf{z} + b )) \label{loss}
\end{equation}

\noindent  where $\sig$ refers to the sigmoid function, and $\mathbf{w}$ are the weights from the dense layer to the output layer and $b$ is a bias term. For multi-class classification the output layer uses a \texttt{softmax} function. Formally, the probability of $k$-th label in the output for classification into $K$ classes:


\begin{equation}
P(y = k|\mathbf{s}, \theta) = \frac{exp~(\mathbf{w}_k^T\mathbf{z} + b_k)} {\sum_{j=1}^{K} exp~({\mathbf{w}_j^T\mathbf{z} + b_j)}} \label{softmax}
\end{equation}

\noindent where, $\mathbf{w}_k$ are the weights associated with class $k$ in the output layer. We fit the models by minimizing the cross-entropy between the predicted distributions $\hat{y}_{n\theta} = p(y_n|\mathbf{s}_n, \theta)$ and the target distributions $y_n$ (i.e., the gold labels).\footnote{Other loss functions (e.g., hinge) yielded similar results.}


\begin{equation}
J (\theta) = \sum_{n=1}^{N} \sum_{k=1}^{K} y_{nk}~log~P(y_n = k|\mathbf{s}_n, \theta) \label{logloss}
\end{equation}

\noindent where, $N$ is the number of training examples and $y_{nk}$ $=$ $I(y_n = k)$ is an indicator variable to encode the gold labels, i.e., $y_{tk}=1$ if the gold label $y_t=k$, otherwise $0$. 




\subsection{Word Embedding and Fine-tuning} \label{sec:word_emb}

We avoid manual engineering of features 
and use 
word embeddings instead as the only features. 
As mentioned before, we can initialize the embeddings $L$ randomly, and learn them as part of model parameters by backpropagating the errors to the look-up layer.
Random initialization may lead the training algorithm to get stuck in local minima. One 
can plug the readily available embeddings from external sources (e.g., Google embeddings \cite{mikolov2013efficient}) in the neural network model and use them as features without further task-specific tuning.
However, this approach does not exploit the automatic feature learning capability of NN models, which is one of the main motivations of using them. In our work, we use the pre-trained word embeddings to better initialize our models, and we fine-tune them for our task
which turns out to be beneficial. More specifically, we initialize the word vectors in $L$ in two different ways. \\


\noindent \textbf{1. Google Embedding:} Mikolov et al. \cite{mikolov2013efficient}  propose two log-linear models for computing word embeddings from large (unlabeled) corpus efficiently: \Ni a \emph{bag-of-words} model CBOW that predicts the current word based on the context words, and \Nii a \emph{skip-gram} model that predicts surrounding words given the current word. 
They released their pre-trained $300$-dimensional word embeddings (vocabulary size $3$ million) trained by the skip-gram model on part of Google news dataset containing about 100 billion words.\footnote{https://code.google.com/p/word2vec/} \\

\noindent \textbf{2. Crisis Embedding:}
Since we work on disaster related tweets, which are quite different from news, we have also trained \emph{domain-specific} embeddings (vocabulary size $20$ million) using the Skip-gram model of \textit{word2vec} tool \cite{mikolov2013distributed} from a large corpus of disaster related tweets. The corpus contains $57,908$ tweets and $9.4$ million tokens. For comparison with Google, we learn word embeddings of $300$-dimensions.

\subsection{Incorporating Other Features} \label{mlp_cnn}

Although CNNs learn word features (i.e., embeddings) automatically, we may still be interested in incorporating other sources of information (e.g., TF-IDF vector representation of tweets) to build a more effective model. Additional features can also guide the training to learn a better model. However, unlike word embeddings, we want these features to be fixed during training. This can be done in our CNN model by creating another channel, which feeds these additional features directly to the \emph{dense} layer. In that case, the dense layer in Equation \ref{dense} can be redefined as

\begin{equation}
\mathbf{z} = f(V' \mathbf{m'} + \mathbf{b_h}) 
\end{equation}

\noindent where $\mathbf{m'}=\mathbf{[m;y]}$ is a concatenated (column) vector of feature maps $\mathbf{m}$ and additional features $\mathbf{y}$, and $V'$ is the associated weight matrix. Notice that by including this additional channel, this network combines a multi-layer perceptron (MLP) with a CNN model.   


\section{Domain Adaptation} \label{sec:adap}

Since DNNs usually have far more parameters than non-neural models, it is often advantageous to train DNNs with lots of data. When comes to big crisis data, one straight-forward approach is to use all available data both from the current event as well as from all past events. However, due to a number of reasons such as, variety of data in different crisis events, different language usage, differences in vocabulary~\cite{chowdhury2013tweet4act}, not all labeled crisis data from past events are useful for classifying event under consideration. In fact in our experiments we found more data but irrelevant degrades the performance of the model. Based on this observation, we apply two domain adaptation techniques that make the best use of the out-of-event data in favor of the event data by either weighing the out-of-event data or intelligently selecting a subset of the out-of-event data. Both methods are described as follows.


\subsection{1. Regularized Adaptation Model:}

Our first approach is to learn an adapted model from the complete data but to 
regularize 
the resultant model towards the event data. Let $\theta_i$ be a CNN model already trained on the event data. We train an adapted model $\theta_a$ on the complete in- and out-of-event data, but regularizing it with respect to the in-domain model $\theta_i$. Formally, we redefine the loss function of Equation \ref{logloss} as follows:  
   
\begin{eqnarray}
J (\theta_a) \hspace{-0.1cm} &=& \hspace{-0.1cm} \sum_{n=1}^{N} \sum_{k=1}^{K} \Bigl[\lambda~y_{nk}~logP(y_n = k|\mathbf{s_n}, \theta_a)+ (1-\lambda) \nonumber \\
&& \hspace{-0.1in}  y_{nk} P(y_n = k|\mathbf{s_n}, \theta_i)~logP(y_n = k|\mathbf{s_n}, \theta_a) \Bigr]  \label {v1}
\end{eqnarray}

\noindent where $P(y_n = k|\mathbf{s_n}, \theta_i)$ is the probability of the training instance $(\mathbf{s_n}, y_n)$ according to the event model $\theta_i$. Notice that the loss function minimizes the cross entropy of the current model $\theta_a$ with respect to the gold labels $y_n$ and the event model $\theta_i$. The mixing parameter $\lambda \in [0,1]$ determines the relative strength of the two components.\footnote{We used a balanced value $\lambda=0.5$ for our experiments.} Similar adaptation model has been  used previously in neural models for machine translation \cite{joty-etAL:2015:EMNLP,durraniEtAl:MT-Summit2015}.

\subsection{2. Instance Selection:}
Similar to the regularized adaptation model, we use an event only classifier to predict the labels of the out-of-event instances. Instead of using probabilities as weights, we select only those out-of-event tweets which are predicted correctly by the event-based classifier.  We concatenate the selected out-of-event tweets to the event data and train a classifier. This method is similar in spirit to the method proposed by Jiang and Zhai~\cite{jiang-zhai:2007}.

\section{Experimental Settings}
\label{sec:settings}
In this section, we first describe the dataset used for the classification tasks. We then present the TF-IDF based features which are used to train the non-neural classification algorithms and used a additional features for the MLP-CNN model. 
In the end, we describe the training settings of non-neural and neural classification models. 

\subsection{Datasets}
We use data from multiple sources: (1) CrisisNLP~\cite{Imran2016CrisisNLP}, (2) CrisisLex~\cite{olteanu2014crisislex}, and (3) AIDR~\cite{imran2014aidr}.
The first two sources have tweets posted during several humanitarian crises and  labeled by paid workers. 
The AIDR data consists of tweets from several crises events labeled by volunteers.\footnote{These are trained volunteers from the Stand-By-Task-Force organization (http://blog.standbytaskforce.com/).} 

The dataset consists of various event types such as earthquakes, floods, typhoons, etc. In all the datasets, the tweets are labeled into various informative classes (e.g., urgent needs, donation offers, infrastructure damage, dead or injured people) and one not-related or irrelevant class. Table \ref{tbl:classes} provides a one line description of each class and also the total number of labels from all the sources. \emph{Other useful information} and \emph{Not related or irrelevant} are the most frequent classes in the dataset. Table \ref{tab:stats} shows statistics about the events we use for our experiments. 

In order to access the difficulty of the classification task, we calculate the inter-annotator agreement (IAA) scores of the datasets obtained from CrisisNLP. The California Earthquake has the highest 
IAA of 0.85 and Typhoon Hagupit has the lowest IAA of 0.70 in the events under-consideration. The IAA of remaining three events are around 0.75. We aim to reach these levels of accuracy. \\
%
%

\begin{table*}[htb]
\centering
\footnotesize
\caption{Description of the classes in the datasets. Column \emph{Labels} shows the total number of annotations for each class}
\label{tbl:classes}
\begin{tabular}{lll}
\toprule
\textbf{Class} & \textbf{Labels} & \textbf{Description}  \\
\midrule
Affected individual  & 6,418 & \begin{tabular}[c]{@{}l@{}}Reports of deaths, injuries, missing, found, or displaced people\end{tabular} \\
\begin{tabular}[c]{@{}l@{}}Donations and volunteering\end{tabular}  & 3,683 & \begin{tabular}[c]{@{}l@{}}Messages containing donations (food, shelter, services etc.) or volunteering offers\end{tabular} \\
\begin{tabular}[c]{@{}l@{}}Infrastructure and utilities\end{tabular} & 3,288 & Reports of infrastructure and utilities damage \\
Sympathy and support & 6,178 & Messages of sympathy-emotional support \\
Other useful information & 14,696 & \begin{tabular}[c]{@{}l@{}}Messages containing useful information that does not fit in one of the above classes\end{tabular} \\
Not related or irrelevant & 15,302 & \begin{tabular}[c]{@{}l@{}}Irrelevant or not informative, or not useful for crisis response\end{tabular}    \\
\bottomrule                                 
\end{tabular}
\end{table*}

\begin{table*}[htb]
\centering
\footnotesize
\begin{tabular}{l |rrrrrr}
\toprule
\textbf{EVENT} & \textbf{Nepal Earthquake} & \textbf{Typhoon Hagupit} & \textbf{California Earthquake}  & \textbf{Cyclone PAM} & \textbf{All Others} \\
\midrule
{\bf Aff}ected individual  & 756 & 204& 227   & 235  & 4624 \\
{\bf Don}ations and volunteering & 1021 & 113 & 83 &  389 & 1752 \\  
{\bf Inf}rastructure and utilities & 351 & 352& 351 & 233 & 1972\\
{\bf Sym}pathy and support  & 983 & 290 & 83 &  164  & 4546\\
{\bf Oth}er Useful Information & 1505 & 732 & 1028  & 679  & 7709\\  
{\bf Not} related or irrelevant  & 6698 & 290& 157 &  718 & 418 \\
\midrule
Grand Total & 11314 & 1981 & 1929 &  2418 & 21021\\
\end{tabular}
\caption{Class distribution of events under consideration and all other crises (i.e. data used as part of {\it out-of-event} data)}
\label{tab:stats}
\end{table*}
\noindent{\bf Data Preprocessing:} We normalize all characters to their lower-cased forms, truncate elongations to two characters, spell out every digit to \texttt{D}, all twitter usernames to \texttt{userID}, and all URLs to \texttt{HTTP}. We remove all punctuation marks except periods, semicolons, question and exclamation marks. We further tokenize the tweets using the CMU TweetNLP tool \cite{gimpel2011part}.\\

\noindent{\bf Data Settings:}
For a particular event such as Nepal earthquake, data from all other events plus \emph{All others} (see Table \ref{tab:stats}) is considered as \emph{out-of-event} data.
We divide each event dataset into train (70\%), validation (10\%) and test sets (20\%) using ski-learn toolkit's module \cite{scikit-learn} which ensured that the class distribution remains reasonably balanced in each subset. \\

\noindent{\bf Feature Extraction:} We extracted unigram, bigram and trigram features from the tweets as features. The features are converted to TF-IDF vectors by considering each tweet as a document. Note that these features are used only in non-neural models. The neural models take tweets and their labels as input. For SVM classifier, we implemented feature selection using Chi Squared test to improve estimator's accuracy scores. 

\subsection{Non-neural Model Settings}
To compare our neural models with the traditional approaches, we experimented with a number of existing models including:  
\Ni Support Vector Machine (\emph{SVM}), a discriminative max-margin model; \Nii Logistic Regression (\emph{LR}), a discriminative probabilistic model; and \Niii Random Forest (\emph{RF}), an ensemble model of decision trees.
We use the implementation from the scikit-learn toolkit \cite{scikit-learn}. All algorithms use the default value of their parameters.

%

\subsection{Settings for Convolutional Neural Network}

We train CNN models by optimizing the cross entropy in Equation \ref{loss} using the gradient-based online learning algorithm ADADELTA \cite{Zeiler12}.\footnote{Other algorithms (SGD, Adagrad) gave similar results.} The learning rate and parameters were set to the values as suggested by the authors. Maximum number of epochs was set to $25$. To avoid overfitting, we use dropout \cite{Srivastava14a} of hidden units and \emph{early stopping} based on the accuracy on the validation set.\footnote{$l_1$ and $l_2$ regularization on weights did not work well.} We experimented with $\{0.0, 0.2, 0.4, 0.5\}$ dropout rates and $\{32, 64, 128\}$ minibatch sizes. 
We limit the vocabulary ($V$) 
to the most frequent $P\%$ ($P\in\{80, 85, 90\}$) words in the training corpus. The word vectors in $L$ were initialized with the pre-trained embeddings. See Section \ref{sec:word_emb}. 
 
We use rectified linear units (ReLU) for the activation functions ($f$), $\{100, 150, 200\}$ filters each having window size ($L$) of $\{2, 3, 4\}$, pooling length ($p$) of $\{2,3, 4\}$, and $\{100, 150, 200\}$ dense layer units. All the hyperparameters are tuned 
on the development set.




\section{Results}
\label{sec:results}
For each event under consideration, we train classifiers on the event data only, on the out-of-event data only, and on a combination of both.
We evaluate them 
for the binary and multi-class classification tasks. For
the former, we merge all informative classes to create one general \emph{Informative} class.

We initialized the CNN model using two types of pre-trained word embeddings. \Ni \emph{Crisis Embeddings\footnote{Using {\it word2vec} with the {\it Skip-gram} method \cite{mikolov2013efficient} and 
{\it context} of size 6} CNN$_{I}$:} trained on all 
crisis tweets data \Nii \emph{Google Embeddings CNN$_{II}$} trained on the Google News dataset. The CNN model then fine-tuned\footnote{We experimented with fixed embeddings but they did not perform well.} the embeddings using the training data.

%


\subsection{Binary Classification}
Table~\ref{tab:binaryresults-auc} (left) presents the results of binary classification comparing several non-neural classifiers 
with 
the CNN-based classifier. 
%
%
CNNs performed better than all non-neural classifiers for all events under consideration.
%
%
\begin{table*}[!htb]
   \caption{\label{tab:binaryresults-auc} (\emph{left table}) The AUC scores of non-neural and neural network-based classifiers. \emph{event}, \emph{out} and \emph{event+out} represents the three different settings of the training data -- event only, out-of-event only and a concatenation of both. (\emph{right table}) Confusion matrix for binary classification of the {\bf Nepal EQ} event. The rows show the actual class as in the gold standard and the column shows the number of tweets predicted in that class.
   }
    \begin{minipage}{.55\linewidth}
      \small
      \centering
       \begin{tabular}{l|ccc|cc}
\toprule
SYS &  RF & LR & SVM & CNN$_{I}$ & CNN$_{II}$\\
\midrule
\multicolumn{6}{c}{Nepal Earthquake} \\
B$_{event}$      & 82.70 & 85.47 & 85.34 & {\bf 86.89} & 85.71 \\
B$_{out}$        & 74.63 & 78.58 & 78.93 & 81.14 & 78.72 \\
B$_{event+out}$  & 81.92 & 82.68 & 83.62 & 84.82 & 84.91 \\
\midrule
 \multicolumn{6}{c}{California Earthquake} \\
B$_{event}$      & 75.64 & 79.57 & 78.95 & {\bf 81.21} & 78.82   \\
B$_{out}$        & 56.12 & 50.37 & 50.83 & 62.08 & 68.82  \\
B$_{event+out}$  & 77.34 & 75.50 & 74.67 & 78.32 & 79.75  \\
\midrule
\multicolumn{6}{c}{Typhoon Hagupit} \\
B$_{event}$      & 82.05 & 82.36 & 78.08 & 87.83 & {\bf 90.17} \\
B$_{out}$        & 73.89 & 71.14 & 71.86 & 82.35 & 84.48 \\
B$_{event+out}$  & 78.37 & 75.90 & 77.64 & 85.84 & 87.71\\
\midrule
\multicolumn{6}{c}{Cyclone PAM} \\
B$_{event}$      & 90.26 & 90.64 & 90.82 & {\bf 94.17} & 93.11   \\
B$_{out}$        & 80.24 & 79.22 & 80.83 & 85.62 & 87.48 	\\
B$_{event+out}$  & 89.38 & 90.61 & 90.74 & 92.64 & 91.20  \\
\bottomrule
\end{tabular}
    \end{minipage}%
    \begin{minipage}{.45\linewidth}
      \centering
      \small
               \begin{tabular}{lcc|cc}
\toprule
&\multicolumn{2}{c}{SVM} & \multicolumn{2}{|c}{CNN$_I$} \\
\midrule
\multicolumn{5}{c}{Event} \\
&Info. & Not Info. & Info. & Not Info. \\
Info. & 639 & 283 & 594 & 328 \\
Not Info. & 179 & 1160 & 128 & 1211 \\
\midrule
 \multicolumn{5}{c}{Out} \\
Info. & 902 & 20 & 635 & 287 \\
Not Info. & 1144 & 195 & 257 & 1082 \\
\midrule
 \multicolumn{5}{c}{Event+out} \\
Info. & 660 & 262 &  606 & 316\\
Not Info. & 263 & 1075  &  174 &1165\\
\midrule
\end{tabular}
    \end{minipage} 
\end{table*}
The improvements are substantial in the case of training with the out-of-event data only. In this case, CNN outperformed SVM by a margin of up to 11\%. 
This result has a significant impact to a situation involving early hours of a crisis, where though a lot of data pours in, 
but performing data labeling using experts or volunteers to get a substantial amount of training data takes a lot of time.
Our result shows that the CNN model handles this situation robustly by making use of the out-of-event data and provides reasonable performance.  

When trained using both the event and out-of-event data, 
CNNs also performed better than the non-neural models. Comparing different training settings, we saw a drop in performance when compared to the event-only training. This is because of the inherent variety in the big crisis data. The large size of the out-of-event data down-weights the benefits of the event data, and skewed the probability distribution
of the training data towards the out-of-event data. 

Table~\ref{tab:binaryresults-auc} (right) presents 
the confusion matrix of the SVM and CNN$_I$ classifiers trained and evaluated on the Nepal earthquake data. The SVM prediction is inclined to
favor the
Informative class whereas CNN predicted more instances as non-informative than informative. In the case of out-of-event training, SVM predicted most of the instances as informative. Thus,
it achieved high recall, but very low precision. CNN, on the other hand, achieved quite balanced precision and recall 
numbers.

To summarize, the neural network based classifier out-performed non-neural classifiers in all data settings. The performance of the models trained on out-of-event data are (as expected) lower than 
that in the other two training settings. However, in case of the CNN models, the results are reasonable to the extent that out-of-event data can be used to predict tweets informativeness when no event data is available. 
Comparing CNN$_{I}$ with CNN$_{II}$, we did not see any system consistently better than
the other.
In the rest of our experiments below, we only consider the CNN$_{I}$ trained on crisis embedding because on the average, the crisis embedding works slightly better than the other alternative.

\subsection{Multi-class Classification}
For the purpose of multi-class classification, we mainly compare the performance of two variations of the CNN-based classifier, CNN$_{I}$ and MLP-CNN$_{I}$ (combining multi-layer perception and CNN), against an SVM classifier.
%
%
\begin{table*}[htb]
\centering
\footnotesize
\begin{tabular}{l|ccc|ccc}
\toprule
SYS & SVM & CNN$_{I}$  & MLP-CNN$_{I}$ & SVM & CNN$_{I}$  & MLP-CNN$_{I}$ \\
    & \multicolumn{3}{c|}{Accuracy} & \multicolumn{3}{c}{Macro F1} \\
\midrule
& \multicolumn{6}{c}{Nepal Earthquake} \\
M$_{event}$      & 70.45  & 72.98 & 73.19 & 0.48 & 0.57 & 0.57\\
M$_{out}$        & 52.81  & 64.88 & 68.46 & 0.46 & 0.51 &  0.51\\
M$_{event+out}$  & 69.61  & 70.80 & 71.47 & 0.55 & 0.55 & 0.55 \\
\midrule
M$_{event+adpt01}$ & 70.00 & 70.50  & 73.10 &  0.56 & 0.55  &  0.56\\
M$_{event+adpt02}$ & 71.20 & 73.15  & {\bf 73.68} &  0.56 & 0.57  & {\bf 0.57}\\
\midrule
& \multicolumn{6}{c}{California Earthquake} \\
M$_{event}$      & 75.66 & 77.80 & 76.85 & 0.65 & 0.70 &  0.70\\
M$_{out}$        & 74.67 & 74.93 & 74.62  & 0.65 & 0.65 & 0.63\\
M$_{event+out}$  & 75.63 & 77.52 & 77.80   & 0.70 & 0.71 & 0.71\\
\midrule
M$_{event+adpt01}$  & 75.60  & 77.32  & 78.52 & 0.68  & 0.71  & 0.70\\
M$_{event+adpt02}$  & 77.32 & 78.52  & {\bf 80.19} & 0.68  & 0.72  & {\bf 0.72}\\
\midrule
& \multicolumn{6}{c}{Typhoon Hagupit} \\
M$_{event}$     & 75.45 & 81.82  & 82.12 & 0.70 & 0.76 & 0.77 \\
M$_{out}$       & 67.64 & 78.79  &  78.18  & 0.63 & 0.75 & 0.73\\
M$_{event+out}$ &  71.10 & 81.51 & 78.81 & 0.68 & 0.79 & 0.78\\
\midrule
M$_{event+adpt01}$ & 72.23  &  81.21  & 81.81 & 0.69  & 0.78  &  0.79\\
M$_{event+adpt02}$ & 76.63  & 83.94  & {\bf 84.24 }& 0.69  & 0.79  & {\bf 0.80}\\
\midrule
& \multicolumn{6}{c}{Cyclone PAM} \\
M$_{event}$     & 68.59 & 70.45 & 71.69 & 0.65 &  0.67 &  0.69\\
M$_{out}$       & 59.58 & 65.70 & 62.19  & 0.57 &  0.63 & 0.59\\
M$_{event+out}$ & 67.88 & 69.01  & 69.21  & 0.63 &  0.65 & 0.66\\
\midrule
M$_{event+adpt01}$ & 67.55 & 71.07 & 72.52  & 0.63  & 0.67  & 0.69\\
M$_{event+adpt02}$ & 68.80 & 71.69 & {\bf 73.35} & 0.66  & 0.69  & {\bf 0.70}\\
\bottomrule
\end{tabular}
\caption{\label{tab:multiclassresults} The accuracy scores of the SVM and CNN based methods with different data settings. 
}
\end{table*}
Table \ref{tab:multiclassresults} summarizes the accuracy and macro-F1 scores. 
In addition to the results on event, out-of-event and event+out, we also show the results for the adaptation methods described in Section \ref{sec:adap}; see the rows with M$_{event+adp01}$ and M$_{event+adpt02}$ system names.  Similar to the results of the binary classification task,  the CNN model outperformed the SVM model in all data settings. 
Combining MLP and CNN improved the  performance of our system significantly. 
The results on training with 
both event and out-of-event data did not have a clear improvement over 
training on the event data only. 
The results dropped slightly in some cases. 

When applied regularized adaptation model (M$_{event+adpt01}$) to adapt the out-of-event data in favor of the event data, the  system shows mixed results. Since the adapted model is using the complete out-of-event data for training, we hypothesized that weighting does not completely balance the effect of the out-of-event data in favor of the event data. In the instance selection adaptation method (M$_{event+adpt02}$), we select items of the out-of-event data that are correctly predicted by the event-based classifier and use them to build an event plus out-of-event classifier. The results improved in all cases.  We expect that a better adaptation method can further improve the classification results which is out of the scope of this work.

\begin{figure*}[htb]
\centering
\includegraphics[height=1.9in]{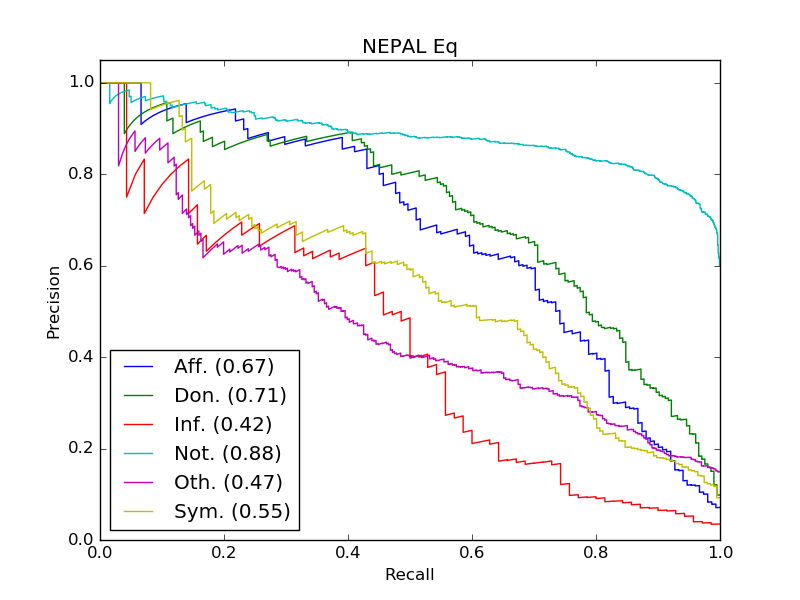}\includegraphics[height=1.9in]{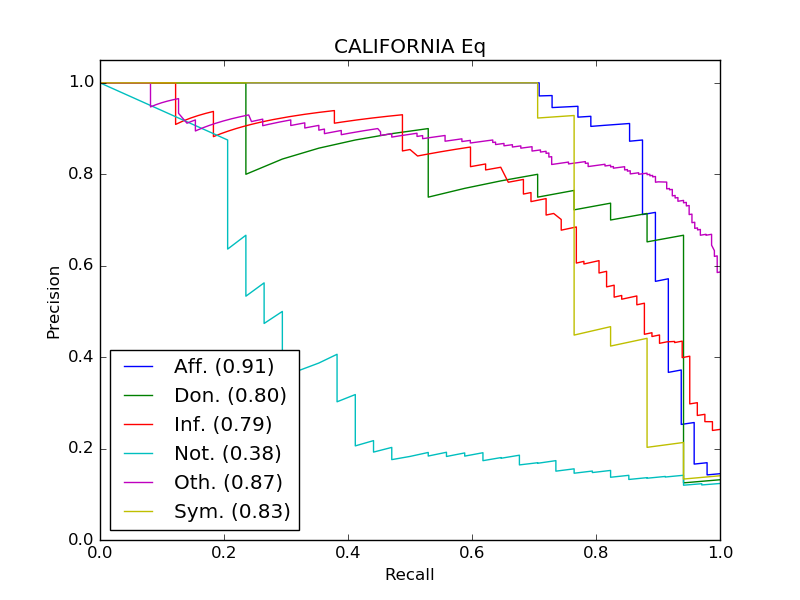}
\includegraphics[height=1.9in]{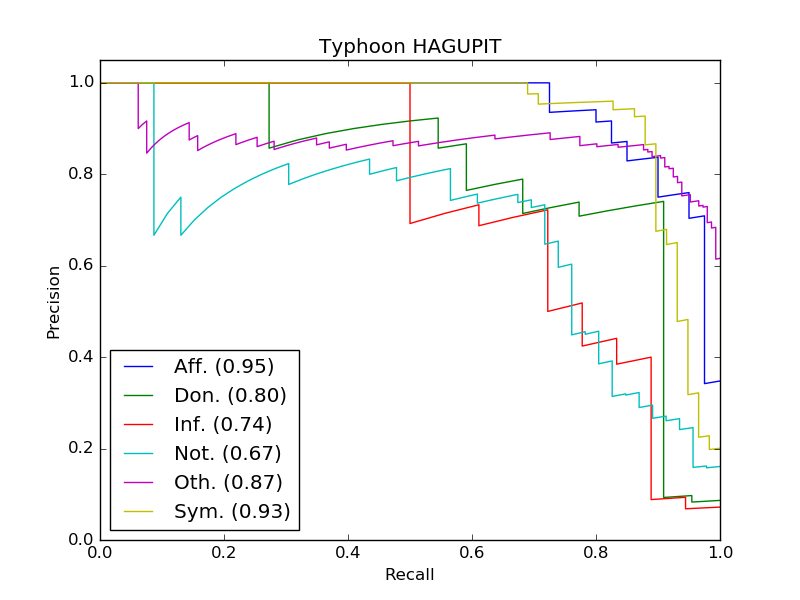}\includegraphics[height=1.9in]{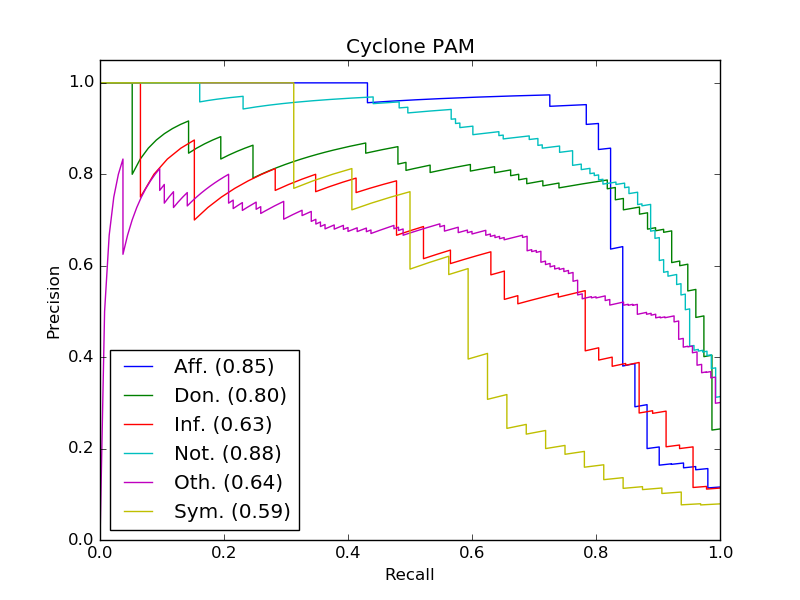}
\caption{\label{fig:precision-curve} Precision-Recall curves and AUC scores of each class generated from MLP-CNN$_{I}$ and M$_{event+adpt02}$ models}
\end{figure*}

\begin{figure}[tb!]
\centering
\includegraphics[height=1.7in]{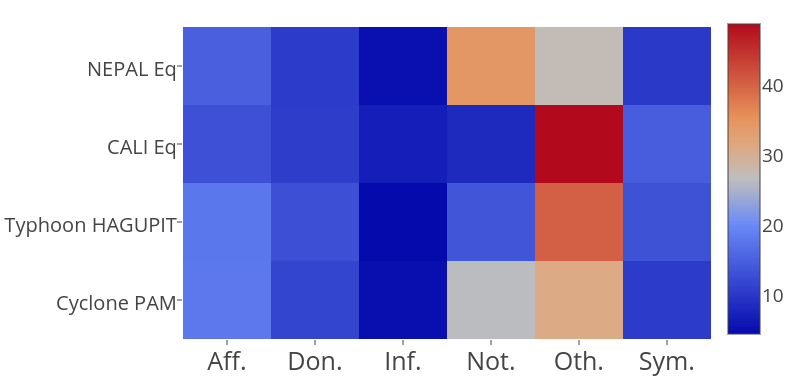}
\caption{\label{fig:heatmap} Heatmap shows class distribution of training data for each event used to train the M$_{event+adapt02}$ models}
\end{figure}

\section{Discussion}
\label{sec:discussion}

Time-critical analysis of big crisis data posted online 
especially during the onset of a crisis situation requires real-time processing capabilities (e.g. real-time classification of messages). 
Big crisis data is voluminous and has a lot of 
variety due to {\it noise}.
Thus, we need to filter out the noise first using a binary (Informative/Non-Informative)
classifier and then use the Informative 
data for humanitarian purposes. 

To fulfill different information needs of humanitarian responders, we have also presented the results obtained from the multi-class classification task. In that case, we used a number of classes useful for humanitarian assistance. Generally the neural network classifiers performed better than the non-neural classifiers. We further showed improvements by doing domain adaptation where 
the instance selection approach outperformed among all neural network classifiers. This is due to the reason that the instance selection helps remove noise from out-of-event labeled data.

In Figure~\ref{fig:precision-curve}, we show precision-recall curves obtained from MLP-CNN$_{I}$ models that outperform
the other models in most of the cases and in Figure~\ref{fig:heatmap} class distribution of the training data used in these models is depicted. 
California, Hagupit, and PAM seem to be easier datasets than Nepal as evidenced by more area under the curve for the former datasets than in Nepal across different classes.
The accuracies of the different classes for the different events seem to have been 
influenced 
somewhat by the number of available training samples.
The areas under the curve for the Nepal earthquake are the lowest implying that it is the hardest event dataset for the classification task.
Interestingly, the Not-Relevant Class is the easiest to identify and has the highest AUC in Nepal (that class has almost 50\% of the tweets for this event) and 
the lowest AUC in the California event datasets (consistent with the fact that that class has less than 10\% of the tweets for this event).
All the other classes have higher AUCs in the California dataset than the Nepal dataset implying California is easier to classify for our classifier.
In Typhoon Hagupit, the not-informative class again has the lowest AUC (about 15\% of the tweets are in this class) and the pattern is similar to that in the California event dataset.  But in Cyclone Pam dataset, 
the Not-Relevant class has greater AUC and the Sympathy (which is the smallest class), Other (which actually has ~30\% support), and Infrastructure classes (around 10\% support) have the lowest AUCs.

Generally, the classes with the smallest percentage of tweets have lower AUCs, but that is not always the case.  We conjecture that a combination of the amount of training data and the inherent hardness related to classifying a class combine to make a class harder or easier to classify. Another factor that we did not consider in this work is the semantic relatedness of a tweet under several classes which brings inconsistency in labeling. Evaluation of the classifiers based on top $N$ probable classes can shed light into this \cite{Magdy2015} where $N$ is the total number of classes in the dataset.


\section{Related Work}
\label{sec:relatedwork}
Studies have analyzed how big crisis data can be useful during major disasters so as to gain insight into the situation as it unfolds~\cite{acar2011twitter,sakaki2010earthquake,varga2013aid}.
A number of systems have been developed to classify, extract, and summarize crisis-relevant information from social media; for a detailed survey see~\cite{imran2015processing}. Cameron, et al., describe a platform for emergency situation awareness~\cite{cameron2012emergency}. 
They classify interesting tweets using an SVM classifier. 
Verma, et al., use Naive Bayes and MaxEnt classifiers to find situational awareness tweets from several crises~\cite{verma2011natural}. 
Imran, et al., implemented AIDR to classify a Twitter data stream during crises~\cite{imran2014aidr}. They use a random forest classifier in an offline setting. 
After receiving every minibatch of 50 training examples, 
they replace the older model with a new one. In~\cite{imran2016cross}, the authors show the performance of a number of non-neural network classifiers trained on labeled data from past crisis events. However, they do not use DNNs in their comparison.
 


DNNs and word embeddings have been applied successfully to address NLP problems~\cite{Weston08deeplearning,collobert2011natural,carageaidentifying}. The emergence of tools such as word2vec \cite{mikolov2013distributed} and GloVe \cite{pennington-socher-manning:2014:EMNLP2014} have enabled NLP researchers to learn word embeddings efficiently and use them to train better models. 

Collobert, et al. \cite{collobert2011natural} presented a unified DNN architecture for solving various NLP tasks including part-of-speech tagging, chunking, named entity recognition and semantic role labeling. They showed that DNNs outperform traditional models in most of these
tasks.  They also proposed a multi-task learning framework for solving the tasks jointly. 

Kim \cite{kim:2014:EMNLP2014} and Kalchbrenner et al. \cite{Kalchbrenner14} used convolutional neural networks (CNN) for sentence-level classification tasks (e.g., sentiment/polarity classification, question classification) and showed that CNNs outperform traditional methods (e.g., SVMs, MaxEnts). Despite these recent advancements, the application of CNNs to disaster response is novel to the best of our knowledge. 

\vspace{-1mm}
\section{Conclusion and Future Work}
\label{sec:conclusion}
\vspace{-1mm}
We addressed the problem of rapid classification of crisis-related data posted on microblogging platforms like Twitter. 
Specifically, we addressed the challenges using deep neural network models for binary and multi-class classification tasks and showed that one can reliably use out-of-event data for the classification of new event when no event-specific data is available. 
The performance of the classifiers degraded from event data when out-of-event training samples were added to training samples.  Thus, we recommend using out-of-event training data during the first few hours of a disaster only after which the training data related to the event should be used.  We improved the performance of the system using domain adaptation by either regularized adaptation and instance selection. 

In the future, we will explore and perform experimentation to determine even more robust domain adaptation techniques.
Moreover, addressing the variety in the data by segregating code mixing of multiple languages in the same tweet needs additional work.  Identifying topic evolution in big crisis data during an event can help us be aware of the variability of such data.  All of these modules can be built on top of our classification system and work in concert with it.

\vspace{-1mm}

\balance
\bibliographystyle{IEEEtran}
\bibliography{IEEEabrv,big_data}

\end{document}